\pdfoutput=1

\documentclass[11pt]{article}

\usepackage[final]{acl}

\usepackage{times}
\usepackage{latexsym}

\usepackage[utf8]{inputenc}

\usepackage{microtype}

\usepackage{inconsolata}

\usepackage{graphicx}
\usepackage{booktabs}
\usepackage{enumitem}
\usepackage{multirow}
\usepackage[greek,english]{babel}
\usepackage{makecell}

\usepackage{textgreek}  
\usepackage{alphabeta}  

\title{Building Multilingual Datasets for Predicting Mental Health Severity through LLMs: Prospects and Challenges}

\author{\textbf{Konstantinos Skianis}$^*$, \textbf{A. Seza Doğruöz}$^\#$, \textbf{John Pavlopoulos}$^\dagger$$^\P$\\
$^*$ Department of Computer Science and Engineering, University of Ioannina, Greece\\
$^\#$ LT3, IDLab, Universiteit Gent, Belgium \\
$^\dagger$ Department of Informatics, Athens University of Economics and Business, Greece\\
$^\P$ Archimedes/AthenaRC, Greece\\
\texttt{kskianis@cse.uoi.gr \quad as.dogruoz@ugent.be \quad annis@aueb.gr}
}

\begin{document}
\maketitle


\begin{abstract}
Large Language Models (LLMs) are increasingly being integrated into various medical fields, including mental health support systems. However, there is a gap in research regarding the effectiveness of LLMs in non-English mental health support applications. To address this problem, we present a novel multilingual adaptation of widely-used mental health datasets, translated from English into six languages (e.g., Greek, Turkish, French, Portuguese, German, and Finnish). This dataset enables a comprehensive evaluation of LLM performance in detecting mental health conditions and assessing their severity across multiple languages. By experimenting with GPT and Llama, we observe considerable variability in performance across languages, despite being evaluated on the same translated dataset. This inconsistency underscores the complexities inherent in multilingual mental health support, where language-specific nuances and mental health data coverage can affect the accuracy of the models.
Through comprehensive error analysis, we emphasize the risks of relying exclusively on LLMs in medical settings (e.g., their potential to contribute to misdiagnoses). Moreover, our proposed approach offers significant cost savings for multilingual tasks, presenting a major advantage for broad-scale implementation.

\end{abstract}

\section{Introduction}

{\it Warning: this paper contains text examples that
are negative, depressive, or adverse.}
\vspace{0.2cm}

Large language models (LLMs) (e.g., GPT-3, BERT, and their multilingual variants) are trained on vast amounts of text data and demonstrate an impressive ability to generate human-like text, perform complex language tasks \cite{devlin2019, brown2020}. 
One of the most promising applications of LLMs is in the healthcare domain, where they have the potential to assist medical specialists with the detection, diagnosis, and treatment of health conditions of patients. 



Mental health disorders (e.g.,  depression, anxiety, and post-traumatic stress disorder (PTSD)) are widespread across the globe, presenting major public health challenges \cite{who2021}. 
Traditional diagnostic approaches typically involve (manual)-reported questionnaires, clinical interviews, and standardized assessments performed by trained professionals \cite{kessler2004}. However, these processes are also time-consuming and labor intensive, leading to delays in treatment and increased waiting times for the patients.

Therefore, leveraging large language models (LLMs) to detect mental health symptoms from textual data has presented a promising alternative. 
These models, for instance, have the ability to process vast amounts of textual content (e.g., social media posts, online forum discussions, personal narratives), and detect specific linguistic indicators related to mental health issues \cite{guntuku2019, chancellor2019}. 
In this way, health professionals can be assisted with the early detection and intervention of mental health symptoms, reduce delays for patients whose symptoms might otherwise go unnoticed. 

However, most studies about applying LLMs to mental health research are still in English and there is a need to explore this research area for other languages \cite{raihan2024mentalhelp}. 
Our study fills these gaps with the following research questions:
\begin{itemize}[noitemsep, topsep=0pt]
\item How accurately can LLMs predict the severity of mental health conditions in English-language through user-generated content?
\item When English-language mental health datasets are translated into other languages using LLMs, does the model maintain similar accuracy in predicting the severity of mental health conditions across these languages?
\end{itemize}
We answer these research questions with the following key contributions: 

\begin{itemize}[noitemsep, topsep=0pt]
\item We create a novel multilingual dataset (covering English, Turkish, French, Portuguese, German, Greek, and Finnish) derived from user-generated social media content, which is automatically translated from English using an LLM.
\footnote{Code and some results are uploaded as supplementary material. The dataset will be made publicly available upon acceptance.}
\item We perform an extensive evaluation of the effectiveness of an LLM in predicting the severity of mental health conditions across these different languages.
\item We provide an in-depth analysis of the model's performance in each language, offering insights into how linguistic diversity affects the task and dataset-specific outcomes.
\end{itemize}


To the best of our knowledge, this is the first study to utilize LLMs for creating and evaluating multilingual adaptations of existing mental health NLP datasets, specifically designed to detect the severity of mental health conditions. Our study pushes forward the creation of inclusive and adaptable AI-powered tools and datasets for mental health diagnosis, emphasizing the critical need for approaches that account for cultural and linguistic diversity in mental health care, especially in non-English contexts with cost-efficient methods.


\section{Related work}

Early investigations into the use of NLP for mental health concentrated on identifying specific symptoms from textual data through machine learning methods.
This often involved analyzing social media content, online health forum discussions, and other text-based sources to find indicators of mental health conditions like depression, anxiety and others \cite{de2014mental, coppersmith2014quantifying, de2014mental,coppersmith2015clpsych, shing2018expert, guntuku2019}. 
Traditional approaches primarily relied on machine learning algorithms with hand-crafted features, which lacked the sophistication of current LLMs.


\paragraph{Multilingual Capabilities of LLMs} 
Although training and evaluation details are not always fully disclosed, LLMs are assumed to be capable of processing and generating text in multiple languages. For instance, the XLM-R model (trained on extensive multilingual data), has demonstrated strong cross-lingual performance. 
As noted by \citet{conneau-etal-2020-unsupervised}, XLM-R surpasses earlier models on various tasks, highlighting the benefits of leveraging large-scale multilingual data.

There are also examples of LLMs excelling in practical applications (e.g., translation and multilingual question answering). For instance, a model created by \citet{xue-etal-2021-mt5}, is pre-trained on a diverse set of languages and demonstrates strong capabilities in translating and answering questions in multiple languages. The model's architecture and training methods allow it to generalize across languages effectively, providing robust performance even in languages with limited training data. 

\begin{figure*}[t]
  \includegraphics[width=\textwidth]{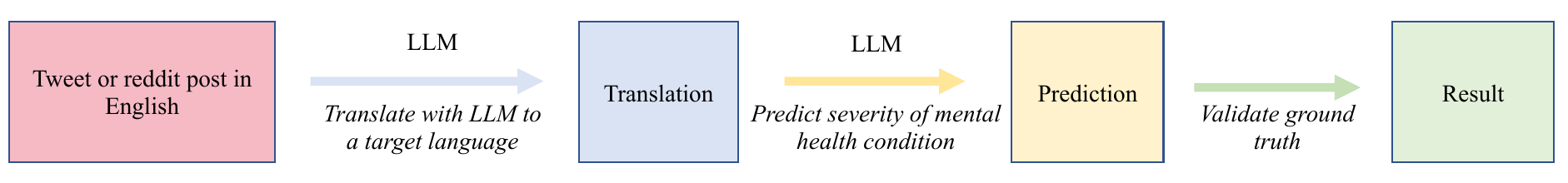} 
  \caption {An illustration of our proposed methodology.}
  \label{proposed}
\end{figure*}

However, major obstacles still hinder the achievement of truly equitable performance across all languages \cite{zhang-etal-2020-improving}. 
One of the primary challenges lies in low-resource languages, which suffer from a scarcity of digitally available text data. 
This often leads to under-performance by LLMs, producing less reliable outputs for such languages. 
Tackling this issue requires not only more inclusive and comprehensive data collection but also innovative approaches in transfer learning that can better leverage sparse resources \cite{dogruoz-sitaram-2022-language}.

Moreover, language is intricately tied to cultural and societal frameworks, making it particularly difficult to capture the subtle, context-driven nuances that vary from one culture to another. This adds another layer of complexity to multilingual NLP tasks.
Addressing these intricacies demands a twofold approach: developing more advanced, culturally-aware algorithms and incorporating datasets that are truly representative of linguistic and cultural diversity \cite{dogruoz-etal-2023-representativeness}. 
Without these efforts, the full potential of multilingual LLMs will remain inaccessible to many, limiting their impact on global applications.

\paragraph{LLMs for Mental Health} LLMs excel at detecting mental health symptoms with greater precision by leveraging their advanced understanding of context and semantics. 
Notable examples include BioBERT \cite{lee2020biobert} and ClinicalBERT \cite{clinicalbert}, which have been pre-trained on biomedical literature and clinical notes. Additionally, models like MentalBERT \cite{ji2022mentalbert}, DisorBERT \cite{aragon2023disorbert}, MentalLLM \cite{xu2024mental} and MentalLlama \cite{yang2023mentalllama} are specifically trained on mental health-related social media data.
Nevertheless, using LLMs for detecting mental health issues may introduce a number of advantages, as well as disadvantages \cite{de2023benefits}.
Additionally, exploiting social network posts represents a promising frontier in both artificial intelligence and mental health care \cite{harrigian2020state}. 
LLMs (e.g., GPT-4 and BERT) can analyze vast amounts of textual data from social media platforms automatically to identify linguistic patterns associated with mental health conditions such as depression, anxiety, and suicidal ideation \cite{ji2023domain}.

Recently, a wide range of social network datasets focusing on mental health have become accessible \cite{raihan2024mentalhelp}. 
The authors compiled posts from platforms like Reddit and Twitter that pertain to conditions such as depression, PTSD, schizophrenia, and eating disorders. 
In addition, they fine-tuned several models using smaller, publicly available annotated mental health datasets to facilitate labeling for their newly introduced MentalHelp dataset. However, the dataset is limited to English-language posts, which constrains its application and hinders further research in multilingual contexts.

\section{Proposed Methodology}
We present a comprehensive methodology for leveraging LLMs and social media posts (in English), to translate them into multiple languages and predict mental health conditions accordingly. 

First, we translate the social media posts across the proposed languages via an LLM. Then, these translations are fed into a prompt that asks the LLM to predict specific severity levels of mental health conditions. Finally, the predicted classes are compared with the ground truth labels and the evaluation metrics are computed.

Our goal is to propose a methodology that can be used to evaluate LLMs on multilingual detection of mental health conditions in languages where minimal relative linguistic resources are publicly available. An illustration of the proposed approach for evaluating LLMs for multilingual detection of mental health conditions is shown in Figure \ref{proposed}.


\begin{table*}[ht]
\centering
  \resizebox{0.9\textwidth}{!}{%
  \begin{tabular}{llcccl}
    \hline
    \textbf{Dataset} & \textbf{Reference} & \textbf{Category} & \textbf{\#Classes} & \textbf{\#Instances} & \textbf{Labels (\#Support)}\\
    \hline
    \multirow{4}{*}{\textsc{Dep-Sev}}& \multirow{4}{*}{\citet{naseem2022early}} & \multirow{4}{*}{Depression} & \multirow{4}{*}{4} & \multirow{4}{*}{3553} & Minimum (2587) \\
    & & & & & Mild (290) \\
    & & & & & Moderate (394) \\
    & & & & & Severe (282) \\
    \midrule
    \multirow{5}{*}{\textsc{C-ssrs}} & 
    \multirow{5}{*}{\citet{gaur2019knowledge}} & \multirow{5}{*}{Suicide} & \multirow{5}{*}{5} & \multirow{5}{*}{500} & Supportive (108)\\
    & & & & & Indicator (99) \\
    & & & & & Ideation (171) \\
    & & & & & Behavior (77) \\
    & & & & & Attempt (45)\\
    \hline
  \end{tabular}
  }
  \caption{Benchmark datasets used in our study.}
  \label{tab:datasets}
\end{table*}

\section{Experiments}


\subsection{Datasets}
In order to test the proposed methodology (above), we use the following datasets:

\paragraph{\textsc{Dep-Severity}} Introduced by \citet{naseem2022early}, this dataset comprises posts from the social media platform Reddit, specifically addressing the varying levels of depression. 
These posts, all in English, are pre-labeled according to four severity levels: minimal, mild, moderate, and severe depression. 
Notably, the dataset is highly imbalanced, with the majority of posts classified under the minimal severity level.

\paragraph{C-SSRS} The C-SSRS (Columbia-Suicide Severity Rating Scale) dataset, created by \citet{gaur2019knowledge}, is a collection of structured data used to assess suicidal ideation and behavior. 
It is based on the Columbia-Suicide Severity Rating Scale, a widely used tool in clinical and research settings to evaluate the severity and frequency of suicidal thoughts, past attempts, and related risk factors. 
The dataset includes several variables (e.g., suicidal ideation intensity, past attempts, preparatory behaviors, and protective factors) allowing researchers and mental health professionals to identify high-risk individuals. \vspace{0.2cm}

\noindent We deliberately chose multi-class datasets to increase the difficulty for the LLM, as binary classification tasks would have been simpler to solve. Table \ref{tab:datasets} provides a summary of key statistics for the two datasets utilized in our study.

\begin{table}[ht]
\centering
  \resizebox{0.9\columnwidth}{!}{%
  \begin{tabular}{lp{5cm}}
    \hline
    \textbf{Dataset} & \textbf{Prediction prompt}\\
    \hline
    \textsc{Dep-Sev} & ``Categorize the following text with 1 of the 4 depression severity levels (0: Minimum, 1: Mild, 2: Moderate, 3: Severe)" \\
    \midrule
    \textsc{C-ssrs} & ``Categorize the content into the following classes based on the user's mental state: 1. Supportive, 2. Indicator, 3. Ideation, 4. Behavior, 5. Attempt."\\
    \hline
  \end{tabular}
  }
  \caption{The used prediction prompts.}
  \label{tab:prompts}
\end{table}

\subsection{Settings}
We use GPT3.5-turbo \cite{brown2020}, GPT-4o-mini via the OpenAI API and Llama3.1 locally to translate the posts and predict the labels. 
Recently introduced models, such as MentaLLaMA \cite{yang2024mentallama}, primarily focus on English, making them unsuitable for use in multilingual settings.
The temperature parameter is set to 0, so the outcome is reproducible (regarding the translations and predictions). We experiment with no examples in the prompt (0-shot) and one example per class (1-shot). 
The task is solved as a classification problem, where we compare the predicted classes with the ground-truth ones, reporting Precision (Pr), Recall (Rec) and F1. We experiment with seven languages from major language families: Turkish (Altaic), French, Portuguese (Romance), English, German (Germanic), Finnish (Uralic), Greek (Hellenic).
Table \ref{tab:prompts} shows the used prompts for predicting the severity levels.


\begin{table*}[t]
\resizebox{\textwidth}{!}{
    \centering
    \begin{tabular}{lccccccccccccccccccccc}
        \toprule
        & \multicolumn{3}{c}{\textbf{English}} & \multicolumn{3}{c}{\textbf{Turkish}} & \multicolumn{3}{c}{\textbf{French}} & \multicolumn{3}{c}{\textbf{Portuguese}} & \multicolumn{3}{c}{\textbf{German}} & \multicolumn{3}{c}{\textbf{Greek}} & \multicolumn{3}{c}{\textbf{Finnish}} \\
        \cmidrule(r){2-4} \cmidrule(r){5-7} \cmidrule(r){8-10} \cmidrule(r){11-13} \cmidrule(r){14-16} \cmidrule(r){17-19} \cmidrule(r){20-22}
        \textbf{Class} & \textbf{Pr} & \textbf{Rec} & \textbf{F1} & \textbf{Pr} & \textbf{Rec} & \textbf{F1} & \textbf{Pr} & \textbf{Rec} & \textbf{F1} & \textbf{Pr} & \textbf{Rec} & \textbf{F1} & \textbf{Pr} & \textbf{Rec} & \textbf{F1} & \textbf{Pr} & \textbf{Rec} & \textbf{F1} & \textbf{Pr} & \textbf{Rec} & \textbf{F1} \\
        \midrule
        \sc minimum & 0.98 & 0.14 & 0.25 & 0.93 & 0.03 & 0.05 & 0.99 & 0.15 & 0.27 & 0.98 & 0.11 & 0.20 & 0.97 & 0.24 & \bf 0.38 & 0.99 & 0.07 & 0.14 & 0.98 & 0.08 & 0.15 \\
        \sc mild      & 0.04 & 0.15 & 0.07 & 0.02 & 0.01 & 0.01 & 0.05 & 0.17 & 0.08 & 0.04 & 0.07 & 0.05 & 0.08 & 0.32 & \bf 0.12 & 0.04 & 0.17 & 0.06 & 0.05 & 0.19 & 0.08 \\
        \sc moderate  & 0.13 & 0.22 & 0.17 & 0.10 & 0.62 & 0.17 & 0.15 & 0.39 & 0.21 & 0.12 & 0.28 & 0.17 & 0.16 & 0.42 & \bf 0.23 & 0.14 & 0.55 & \bf 0.23 & 0.12 & 0.31 & 0.17 \\
        \sc severe    & 0.13 & 0.71 & 0.22 & 0.13 & 0.39 & 0.19 & 0.15 & 0.58 & \bf 0.23 & 0.12 & 0.74 & 0.20 & 0.16 & 0.35 & 0.22 & 0.16 & 0.28 & 0.20 & 0.12 & 0.55 & 0.20 \\
        \midrule
        \textbf{Macro avg} & 0.32 & 0.30 & 0.17 & 0.29 & 0.26 & 0.11 & 0.33 & 0.32 & 0.20 & 0.31 & 0.30 & 0.16 & 0.34 & 0.33 & \underline{\bf 0.24} & 0.33 & 0.27 & 0.16 & 0.32 & 0.28 & 0.15 \\
        \bottomrule
    \end{tabular}
    }
    \caption{\textbf{GPT-3.5 with 0-shot learning on \textsc{Dep-Severity}}, measuring Precision, Recall, and F1 per class per language. The last row shows the macro averages. The best F1 per class is shown in bold. The best macro-averaged F1 across languages is underlined.}
    \label{results-depression}
\end{table*}

\begin{table*}[t] 
\resizebox{\textwidth}{!}{ 
\centering \begin{tabular}{lccccccccccccccccccccc} \toprule & \multicolumn{3}{c}{\textbf{English}} & \multicolumn{3}{c}{\textbf{Turkish}} & \multicolumn{3}{c}{\textbf{French}} & \multicolumn{3}{c}{\textbf{Portuguese}} & \multicolumn{3}{c}{\textbf{German}} & \multicolumn{3}{c}{\textbf{Greek}} & \multicolumn{3}{c}{\textbf{Finnish}} \\ \cmidrule(r){2-4} \cmidrule(r){5-7} \cmidrule(r){8-10} \cmidrule(r){11-13} \cmidrule(r){14-16} \cmidrule(r){17-19} \cmidrule(r){20-22} \textbf{Class} & \textbf{Pr} & \textbf{Rec} & \textbf{F1} & \textbf{Pr} & \textbf{Rec} & \textbf{F1} & \textbf{Pr} & \textbf{Rec} & \textbf{F1} & \textbf{Pr} & \textbf{Rec} & \textbf{F1} & \textbf{Pr} & \textbf{Rec} & \textbf{F1} & \textbf{Pr} & \textbf{Rec} & \textbf{F1} & \textbf{Pr} & \textbf{Rec} & \textbf{F1} \\ 
\midrule 
\sc Supportive & 0.27 & 0.13 & 0.18 & 0.40 & 0.02 & 0.04 & 0.30 & 0.19 & 0.24 & 0.36 & 0.04 & 0.07 & 0.38 & 0.27 & 0.31 & 0.39 & 0.35 & \bf 0.37 & 0.29 & 0.09 & 0.14 \\ 
\sc Indicator & 0.21 & 0.76 & 0.33 & 0.20 & 0.95 & 0.34 & 0.21 & 0.72 & 0.32 & 0.21 & 0.92 & 0.34 & 0.27 & 0.57 & \bf 0.37 & 0.22 & 0.49 & 0.31 & 0.22 & 0.44 & 0.30 \\ 
\sc Ideation & 0.45 & 0.26 & 0.33 & 0.25 & 0.05 & 0.08 & 0.40 & 0.21 & 0.28 & 0.30 & 0.09 & 0.14 & 0.40 & 0.49 & \bf 0.44 & 0.39 & 0.42 & 0.40 & 0.36 & 0.56 & 0.44 \\ 
\sc Behavior & 0.00 & 0.00 & 0.00 & 0.00 & 0.00 & 0.00 & 0.00 & 0.00 & 0.00 & 0.00 & 0.00 & 0.00 & 0.50 & 0.03 & \bf 0.05 & 0.00 & 0.00 & 0.00 & 0.00 & 0.00 & 0.00 \\ 
\sc Attempt & 0.00 & 0.00 & 0.00 & 0.25 & 0.02 & 0.04 & 0.00 & 0.00 & 0.00 & 1.00 & 0.02 & 0.04 & 1.00 & 0.04 & \bf 0.09 & 0.00 & 0.00 & 0.00 & 0.00 & 0.00 & 0.00 \\ 
\midrule \textbf{Macro avg} & 0.19 & 0.23 & 0.17 & 0.22 & 0.21 & 0.10 & 0.18 & 0.22 & 0.17 & 0.37 & 0.21 & 0.12 & 0.51 & 0.28 & \underline{\bf 0.25} & 0.20 & 0.25 & 0.22 & 0.17 & 0.22 & 0.18 \\ 
\bottomrule 
\end{tabular} 
} 
\caption{\textbf{GPT-3.5-turbo with 0-shot learning on \textsc{C-ssrs}}, measuring Precision, Recall, and F1 per class per language. The last row shows the macro averages. The best F1 per class is shown in bold. The best average F1 across languages is underlined.} 
\label{results-suicide-gpt3.5} 
\end{table*}

\begin{table*}[t] 
\resizebox{\textwidth}{!}{ 
\centering \begin{tabular}{lccccccccccccccccccccc} \toprule & \multicolumn{3}{c}{\textbf{English}} & \multicolumn{3}{c}{\textbf{Turkish}} & \multicolumn{3}{c}{\textbf{French}} & \multicolumn{3}{c}{\textbf{Portuguese}} & \multicolumn{3}{c}{\textbf{German}} & \multicolumn{3}{c}{\textbf{Greek}} & \multicolumn{3}{c}{\textbf{Finnish}} \\ \cmidrule(r){2-4} \cmidrule(r){5-7} \cmidrule(r){8-10} \cmidrule(r){11-13} \cmidrule(r){14-16} \cmidrule(r){17-19} \cmidrule(r){20-22} \textbf{Class} & \textbf{Pr} & \textbf{Rec} & \textbf{F1} & \textbf{Pr} & \textbf{Rec} & \textbf{F1} & \textbf{Pr} & \textbf{Rec} & \textbf{F1} & \textbf{Pr} & \textbf{Rec} & \textbf{F1} & \textbf{Pr} & \textbf{Rec} & \textbf{F1} & \textbf{Pr} & \textbf{Rec} & \textbf{F1} & \textbf{Pr} & \textbf{Rec} & \textbf{F1} \\ 
\midrule \sc minimum & 0.97 & 0.32 & 0.48 & 0.97 & 0.18 & 0.30 & 0.97 & 0.30 & 0.46 & 0.97 & 0.31 & 0.47 & 0.96 & 0.36 & \bf 0.52 & 0.98 & 0.26 & 0.41 & 0.97 & 0.26 & 0.40 \\ 
\sc mild & 0.07 & 0.26 & \bf 0.11 & 0.04 & 0.18 & 0.07 & 0.05 & 0.14 & 0.07 & 0.07 & 0.22 & 0.11 & 0.06 & 0.14 & 0.08 & 0.05 & 0.13 & 0.07 & 0.05 & 0.17 & 0.07 \\ 
\sc moderate & 0.18 & 0.41 & \bf 0.25 & 0.15 & 0.42 & 0.23 & 0.16 & 0.50 & 0.24 & 0.16 & 0.41 & 0.23 & 0.15 & 0.42 & 0.22 & 0.15 & 0.52 & 0.23 & 0.15 & 0.48 & 0.23 \\ 
\sc severe & 0.19 & 0.43 & \bf 0.26 & 0.17 & 0.48 & 0.25 & 0.18 & 0.44 & \bf 0.26 & 0.17 & 0.49 & 0.25 & 0.18 & 0.48 & \bf 0.26 & 0.17 & 0.40 & 0.24 & 0.18 & 0.40 & 0.25 \\ 
\midrule \textbf{Macro avg} & 0.35 & 0.35 & \underline{\bf 0.27} & 0.33 & 0.31 & 0.21 & 0.34 & 0.34 & 0.26 & 0.34 & 0.36 & \underline{\bf 0.27} & 0.34 & 0.35 & \underline{\bf 0.27} & 0.34 & 0.33 & 0.24 & 0.34 & 0.33 & 0.24 \\ \bottomrule \end{tabular} 
} 
\caption{\textbf{GPT-4o-mini with 0-shot learning on \textsc{Dep-Severity}}, measuring Precision, Recall, and F1 per class per language. The last row shows the macro averages. The best F1 per class is shown in bold. The best macro-averaged F1 across languages is underlined.} 
\label{results-depression-gpt4} 
\end{table*}

\begin{table*}[t] 
\resizebox{\textwidth}{!}{ 
\centering \begin{tabular}{lccccccccccccccccccccc} \toprule & \multicolumn{3}{c}{\textbf{English}} & \multicolumn{3}{c}{\textbf{Turkish}} & \multicolumn{3}{c}{\textbf{French}} & \multicolumn{3}{c}{\textbf{Portuguese}} & \multicolumn{3}{c}{\textbf{German}} & \multicolumn{3}{c}{\textbf{Greek}} & \multicolumn{3}{c}{\textbf{Finnish}} \\ \cmidrule(r){2-4} \cmidrule(r){5-7} \cmidrule(r){8-10} \cmidrule(r){11-13} \cmidrule(r){14-16} \cmidrule(r){17-19} \cmidrule(r){20-22} \textbf{Class} & \textbf{Pr} & \textbf{Rec} & \textbf{F1} & \textbf{Pr} & \textbf{Rec} & \textbf{F1} & \textbf{Pr} & \textbf{Rec} & \textbf{F1} & \textbf{Pr} & \textbf{Rec} & \textbf{F1} & \textbf{Pr} & \textbf{Rec} & \textbf{F1} & \textbf{Pr} & \textbf{Rec} & \textbf{F1} & \textbf{Pr} & \textbf{Rec} & \textbf{F1} \\ 
\midrule 
\sc Supportive & 0.40 & 0.65 & 0.50 & 0.40 & 0.69 & 0.51 & 0.43 & 0.60 & 0.50 & 0.49 & 0.57 & \bf 0.53 & 0.44 & 0.56 & 0.50 & 0.45 & 0.54 & 0.49 & 0.39 & 0.70 & 0.50 \\ 
\sc Indicator & 0.26 & 0.52 & 0.34 & 0.27 & 0.36 & 0.31 & 0.27 & 0.44 & 0.33 & 0.29 & 0.55 & \bf 0.38 & 0.24 & 0.46 & 0.32 & 0.27 & 0.53 & 0.36 & 0.21 & 0.29 & 0.25 \\ 
\sc Ideation & 0.48 & 0.20 & 0.29 & 0.46 & 0.35 & 0.40 & 0.46 & 0.32 & 0.37 & 0.43 & 0.26 & 0.33 & 0.42 & 0.36 & 0.39 & 0.46 & 0.46 & \bf 0.46 & 0.44 & 0.37 & 0.40 \\ 
\sc Behavior & 1.00 & 0.01 & 0.03 & 0.31 & 0.06 & \bf 0.11 & 0.67 & 0.03 & 0.05 & 0.50 & 0.01 & 0.03 & 0.00 & 0.00 & 0.00 & 0.00 & 0.00 & 0.00 & 0.50 & 0.01 & 0.03 \\ 
\sc Attempt & 0.39 & 0.47 & 0.42 & 0.30 & 0.22 & 0.26 & 0.31 & 0.44 & 0.37 & 0.34 & 0.62 & \bf 0.44 & 0.42 & 0.22 & 0.29 & 0.57 & 0.09 & 0.15 & 0.35 & 0.13 & 0.19 \\ 
\midrule \textbf{Macro avg} & 0.51 & 0.37 & 0.32 & 0.35 & 0.34 & 0.32 & 0.43 & 0.37 & 0.33 & 0.41 & 0.40 & \underline{\bf 0.34} & 0.30 & 0.32 & 0.30 & 0.35 & 0.32 & 0.29 & 0.38 & 0.30 & 0.27 \\ 
\bottomrule \end{tabular} } \caption{\textbf{GPT-4o-mini with 0-shot learning on \textsc{C-ssrs}}, measuring Precision, Recall, and F1 per class per language. The last row shows the macro averages. The best F1 per class is shown in bold. The best average F1 across languages is underlined.} 
\label{results-suicide-gpt-4o-mini} 
\end{table*}

\begin{table}[t] 
\resizebox{\columnwidth}{!}{ 
\centering \begin{tabular}{lcccccccccccc} \toprule & \multicolumn{3}{c}{\textbf{English}} & \multicolumn{3}{c}{\textbf{French}} & \multicolumn{3}{c}{\textbf{Portuguese}} & \multicolumn{3}{c}{\textbf{German}} \\ \cmidrule(r){2-4} \cmidrule(r){5-7} \cmidrule(r){8-10} \cmidrule(r){11-13} \textbf{Class} & \textbf{Pr} & \textbf{Rec} & \textbf{F1} & \textbf{Pr} & \textbf{Rec} & \textbf{F1} & \textbf{Pr} & \textbf{Rec} & \textbf{F1} & \textbf{Pr} & \textbf{Rec} & \textbf{F1} \\ 
\midrule \sc minimum & 0.88 & 0.12 & 0.22 & 0.91 & 0.23 & 0.37 & 0.95 & 0.18 & 0.30 & 0.95 & 0.25 & \bf 0.40 \\ 
\sc mild & 0.04 & 0.20 & 0.07 & 0.04 & 0.13 & 0.07 & 0.05 & 0.14 & \bf 0.08 & 0.05 & 0.15 & \bf 0.08 \\ 
\sc moderate & 0.15 & 0.60 & \bf 0.25 & 0.16 & 0.58 & 0.24 & 0.14 & 0.76 & 0.24 & 0.14 & 0.41 & 0.21 \\ 
\sc severe & 0.18 & 0.22 & \bf 0.20 & 0.12 & 0.26 & 0.17 & 0.20 & 0.11 & 0.14 & 0.13 & 0.41 & \bf 0.20 \\ 
\midrule \textbf{Macro avg} & 0.32 & 0.29 & 0.19 & 0.31 & 0.30 & 0.21 & 0.34 & 0.30 & 0.19 & 0.32 & 0.31 & \underline{\bf 0.22} \\ 
\bottomrule 
\end{tabular} 
} 
\caption{\textbf{Llama3.1 with 0-shot learning on \textsc{Dep-Severity}}, measuring Precision, Recall, and F1 per class per language. The last row shows the macro averages. The best F1 per class is shown in bold. The best average F1 across languages is underlined.} 
\label{results-depression-llama} 
\end{table}

\begin{table*}[t] 
\resizebox{\textwidth}{!}{ 
\centering \begin{tabular}{lccccccccccccccccccccc} \toprule & \multicolumn{3}{c}{\textbf{English}} & \multicolumn{3}{c}{\textbf{Turkish}} & \multicolumn{3}{c}{\textbf{French}} & \multicolumn{3}{c}{\textbf{Portuguese}} & \multicolumn{3}{c}{\textbf{German}} & \multicolumn{3}{c}{\textbf{Greek}} & \multicolumn{3}{c}{\textbf{Finnish}} \\ \cmidrule(r){2-4} \cmidrule(r){5-7} \cmidrule(r){8-10} \cmidrule(r){11-13} \cmidrule(r){14-16} \cmidrule(r){17-19} \cmidrule(r){20-22} \textbf{Class} & \textbf{Pr} & \textbf{Rec} & \textbf{F1} & \textbf{Pr} & \textbf{Rec} & \textbf{F1} & \textbf{Pr} & \textbf{Rec} & \textbf{F1} & \textbf{Pr} & \textbf{Rec} & \textbf{F1} & \textbf{Pr} & \textbf{Rec} & \textbf{F1} & \textbf{Pr} & \textbf{Rec} & \textbf{F1} & \textbf{Pr} & \textbf{Rec} & \textbf{F1} \\ \midrule \sc minimum & 0.90 & 0.14 & 0.25 & 0.76 & 0.23 & 0.36 & 0.91 & 0.11 & 0.19 & 0.90 & 0.11 & 0.20 & 0.76 & 0.75 & \bf 0.76 & 0.79 & 0.02 & 0.05 & 0.68 & 0.16 & 0.25 \\ 
\sc mild & 0.03 & 0.06 & 0.04 & 0.07 & 0.10 & \bf 0.08 & 0.04 & 0.04 & 0.04 & 0.05 & 0.14 & 0.07 & 0.07 & 0.08 & 0.07 & 0.05 & 0.10 & 0.06 & 0.03 & 0.08 & 0.05 \\ 
\sc moderate & 0.14 & 0.65 & \bf 0.23 & 0.11 & 0.29 & 0.16 & 0.13 & 0.44 & 0.20 & 0.15 & 0.51 & \bf 0.23 & 0.16 & 0.23 & 0.19 & 0.14 & 0.26 & 0.18 & 0.13 & 0.39 & 0.19 \\
\sc severe & 0.15 & 0.35 & \bf 0.21 & 0.09 & 0.40 & 0.14 & 0.11 & 0.61 & 0.19 & 0.13 & 0.48 & 0.20 & 0.16 & 0.07 & 0.09 & 0.10 & 0.73 & 0.17 & 0.10 & 0.37 & 0.16 \\ 
\midrule \textbf{Macro avg} & 0.30 & 0.30 & 0.18 & 0.26 & 0.26 & 0.18 & 0.30 & 0.30 & 0.16 & 0.31 & 0.31 & 0.18 & 0.29 & 0.28 & \underline{\bf 0.28} & 0.27 & 0.28 & 0.12 & 0.23 & 0.25 & 0.16 \\ 
\bottomrule 
\end{tabular} 
} 
\caption{\textbf{GPT-3.5 with 1-shot learning on \textsc{Dep-Severity}} by measuring Precision, Recall, and F1 per class per language. The last row shows macro averages for precision, recall, and F1. Best F1 per class is shown in bold. Best average F1 across languages is underlined.} \label{results-depression-gpt3.5-add-shot} 
\end{table*}

\subsection{Preliminary Prompting} 
Before presenting any posts to our LLM, whether for translation or classification, we first inquired how it would approach categorizing them into varying levels of depression severity. The LLM explained that it would begin by identifying specific language cues linked to depressive states, such as:

\begin{itemize} \setlength\itemsep{0.01cm} 
\item Recurrent negative emotions like sadness or a sense of hopelessness. 
\item Self-critical thoughts or feelings of worthlessness. 
\item Expressions reflecting isolation or withdrawal from social interactions. \item Observable shifts in behavior or routines (e.g., disrupted sleep or changes in appetite). 
\item Mentions of emotional suffering or psychological distress. 
\end{itemize}

Furthermore, the LLM would attempt to adapt these signals to the four depression severity levels in the context of social media content. For instance:
\begin{itemize}
\setlength\itemsep{0.01cm} 
\item Level 1 (Minimal): Posts that show infrequent or minor expressions of sadness or frustration. 
\item Level 2 (Mild): Posts where negative emotions or notable behavioral shifts are more consistent. 
\item Level 3 (Moderate): Posts indicating substantial interference with daily life or clear markers of emotional distress. 
\item Level 4 (Severe): Posts showing extreme emotional distress, risk of self-harm, or complete social disengagement. 
\end{itemize}
From this, we can infer that the LLM anticipates posts containing broad indicators of depressive symptoms.

Similarly, for the second dataset, we expect the LLM to detect various linguistic and contextual features in the social media posts to classify them.
Some key aspects that it considers per class are:

\begin{enumerate}

\item Supportive (Encouragement and Positive Reinforcement). Some key linguistic features that may appear include expressions of empathy, encouragement, positivity and use of supportive phrases (e.g., "You are not alone", "Stay strong", "I'm here for you", "You matter"). 


\item Indication (Signs of Distress or Risk). Linguistic features may include the use of worrying expressions (e.g., "I feel lost", "I don't know what to do", "Nothing makes sense anymore", "I'm so tired"; indirect mentions of mental health struggles: "I'm always exhausted", "Everything is meaningless", "I just want it to stop"). 

\item Ideation (Expressions of Suicidal Thoughts). Key linguistic features may include direct expressions of suicidal thoughts (e.g., "I wish I didn’t exist", "I want to end it all", "I don’t see the point of living"; questions about death and suicide: "What happens after we die?", "Is there peace after death?"; use of hopelessness in an extreme form: "Nothing will ever change", "I just want to disappear"). 


\item Behavior (Actions Related to Suicide). Key linguistic features include descriptions of planning or preparation (e.g., "I've been researching painless ways to go", "I wrote my letter today", "This might be my last post", "I have the pills ready"; mentions of giving away possessions or saying goodbye: "I just want to thank everyone for being in my life"). 


\item Attempt (Reports of Actual Suicide Attempts). Key linguistic features include explicit mentions of self-harm or suicide attempts (e.g., "I tried last night but failed", "I was in the hospital after overdosing", "I attempted but someone stopped me", "I survived but I don't know if I should be happy about it") and descriptions of medical intervention (e.g., "They pumped my stomach", "They found me just in time", "I'm in the ER right now"). 


\end{enumerate}


\subsection{Experimental results} 
The results are organized and presented in the following subsections: classification in the source language, classification in the target language, classification across languages, and further analysis of LLMs performance with multilingual examples. 

\paragraph{Source language}\label{sssec:results_source}
Initially, we experimented with the data in their source language (English), to set the baseline performance. 
In other words, no translation step has been performed at this stage. When using GPT-3.5, the first column in Tables~\ref{results-depression} and \ref{results-suicide-gpt3.5} presents the results on depression and suicide detection tasks respectively. 
The best F1 in both tasks is achieved for the lowest severity/indication (minimum depression and supportive for suicide). 
In \textsc{Dep-Severity}, the best performance is achieved for the two edges of lowest (F1=0.48) and highest severity (F1=0.26). 
In \textsc{C-ssrs}, on the other hand, F1 drops from 0.18 to 0.00. 
When using the superior GPT-4o-mini (Tables \ref{results-depression-gpt4} and \ref{results-suicide-gpt-4o-mini} ), F1 increases to 0.27 and to 0.32, respectively. 
Detecting specific levels of suicidal ideation is harder than in the case of depression, as they may be considered less distinct. Therefore, it could be probably more challenging for the LLM to distinguish between them, also shown by the results for some classes.

\paragraph{Target languages}

\textbf{GPT-3.5} has a varying performance when operating on content translated from English to target languages. 
In \textsc{Dep-Severity} (Table~\ref{results-depression}), performance drops in Turkish (-6), Portuguese (-1), Greek (-1), and Finish (-2). 
On the other hand, the performance increases in French (+3) and German (+7). 
In \textsc{C-ssrs} (Table~\ref{results-suicide-gpt3.5}), performance drops in all target languages, with the drop for Turkish (-7) being the most considerable one. 
This can be explained by the limited number of Turkish resources in health, especially in the domain of mental health \cite{ccoltekin2023resources}. 

\noindent\textbf{GPT-4o-mini} in \textsc{Dep-Severity} (Table~\ref{results-depression-gpt4}) performs equally well in Portuguese and German (as in English). However, this is not true for the rest of the target languages. 
In \textsc{C-ssrs}, performance remains similar for most target languages. 
However, we observe again the phenomenon of better performance for a target language, namely for French (+1) and Portuguese (+2), as was the case for GPT-3.5 in German. 
In general, GPT-4o-mini is much more stable and consistent for correct prediction.

\begin{figure}[h]\centering
  \includegraphics[width=0.8\columnwidth]{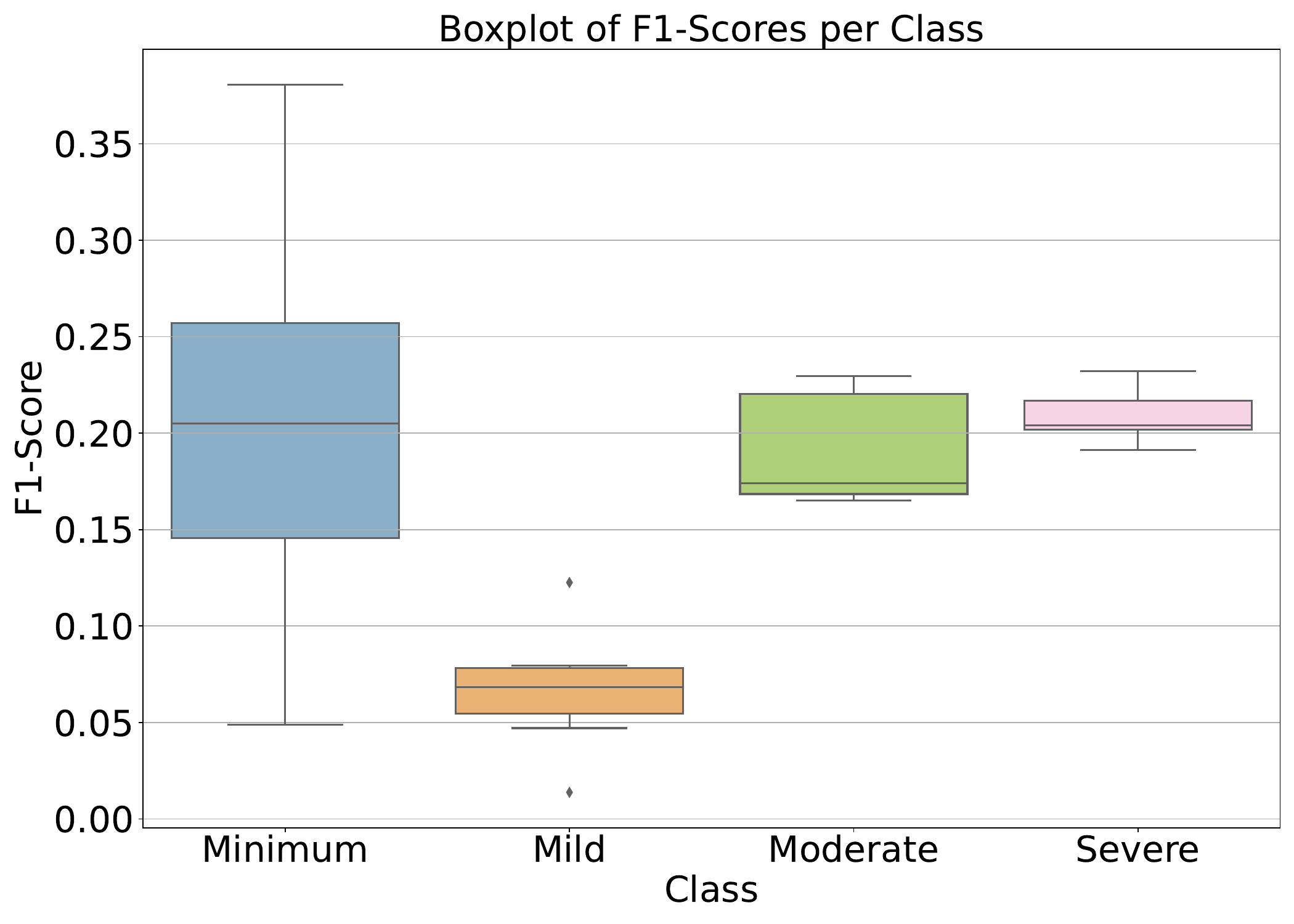}
  \caption {F1 of GPT-3.5 across languages (vertically) per class (box) for \textsc{Dep-Severity}.}  
  \label{boxplot-depression}
\end{figure}
\begin{figure}[t]\centering
  \includegraphics[width=0.8\columnwidth]{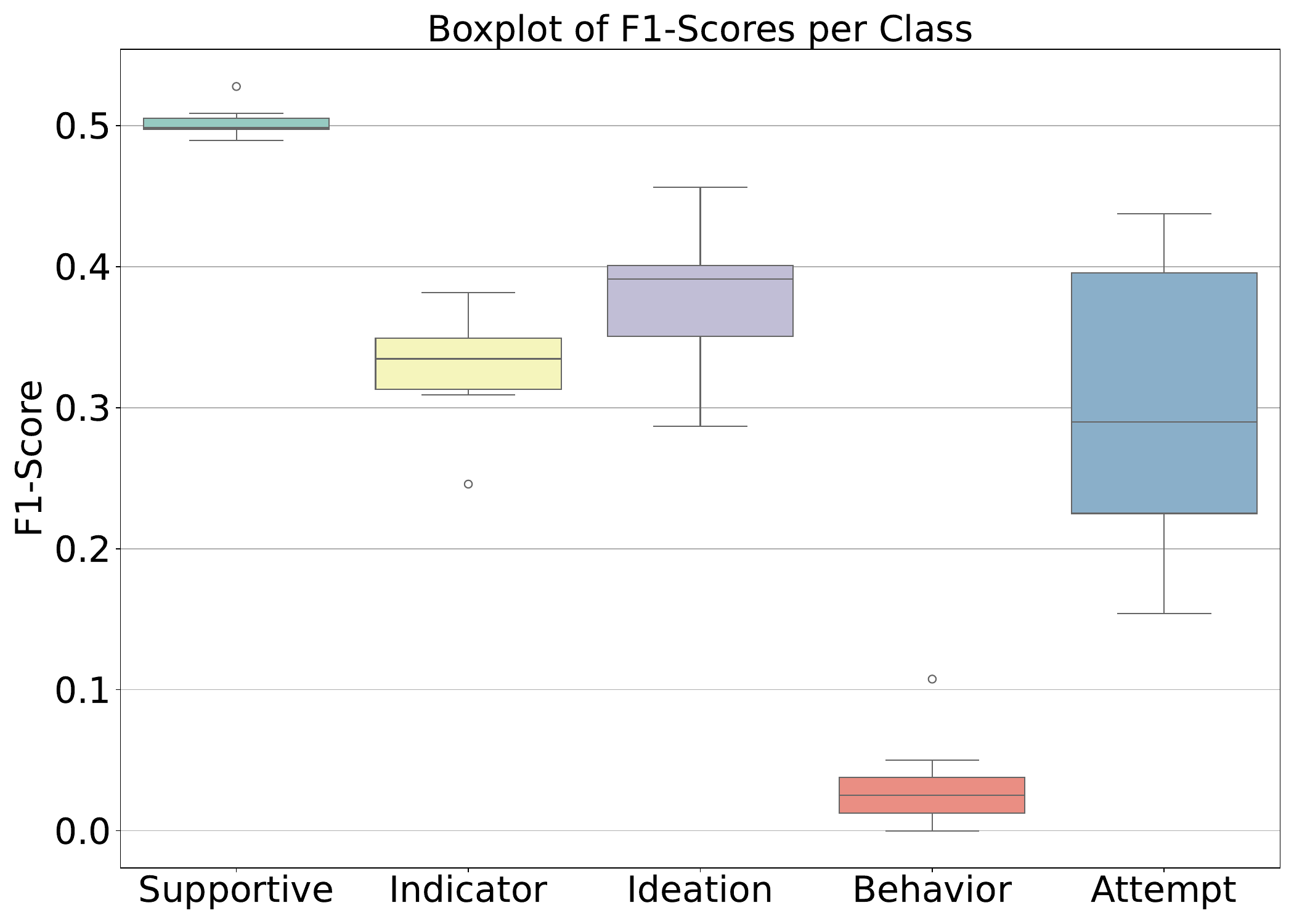}
  \caption {F1 of GPT-4o-mini across languages (vertically) per class (box) for \textsc{C-ssrs}.}
  \label{boxplot-suicide}
\end{figure}

\paragraph{Classification across languages}
Figures~\ref{boxplot-depression} and~\ref{boxplot-suicide} present F1 in boxplots, summarising the performance of GPT-3.5 across languages per task. 
In \textsc{Dep-Severity}, the lowest (minimum) severity class is characterized by a wide range, from 0.15 to 0.25. The whiskers are also wide, from  0.05 (Turkish) to 0.38 (German). Mild severity is mishandled across languages while for the classes of moderate and severe, the model achieves a relatively high score across languages. 
In \textsc{C-ssrs}, the best performance is achieved for the ``Supportive" category. 
This is also the second major class, while slightly worse performance is achieved for the other classes. 
For all classes, apart from ``Attempt" the performance is similarly distributed across languages. 

\paragraph{From zero to single shot}
Table~\ref{results-depression-gpt3.5-add-shot} presents the results when 1-shot learning is employed. In English, performance remains similarly low in \textsc{Dep-Severity}. 
Again, performance is better for a target language (German), compared to the performance for the source language. 
In general, we observe an increase going from 0-shot to 1-shot.
Better performance could be observed if more examples were provided. However, this would also increase the cost of the experiments by approx. doubling the size of the prompt.

\paragraph{Experimenting with Llama-3.1} 
Table~\ref{results-depression-llama} shows the performance of Llama-3.1 in \textsc{Dep-Severity} using 0-shot, for the languages it was available for. Although the model performs better in English compared to GPT-3.5 (both 0- and 1-shot), it is slightly worse in target languages. 
Notably, the best performance is not achieved in English.

\paragraph{Back-translation}
To assess the effect of translation, we translated the already translated texts back to English and repeated the experiments for GPT-3.5-turbo with 0-shot on \textsc{C-ssrs}. Table~\ref{results-suicide-gpt-3.5-back} shows that the prediction increased when translating back to English. The highest gain (from 0.17; see Table~\ref{results-suicide-gpt3.5}) is observed when translating to Portuguese or Finnish and back (+0.05) while +0.03 is reported for Turkish and +0.02 for the rest. Hence, we observe that translating before predicting assists the prediction performance of the LLM. 
We present some representative examples in the Appendix.

\begin{table*}[t] 
\resizebox{\textwidth}{!}{ 
\centering \begin{tabular}{lccccccccccccccccccc} \toprule & \multicolumn{3}{c}{\textbf{Turkish}} & \multicolumn{3}{c}{\textbf{French}} & \multicolumn{3}{c}{\textbf{Portuguese}} & \multicolumn{3}{c}{\textbf{German}} & \multicolumn{3}{c}{\textbf{Greek}} & \multicolumn{3}{c}{\textbf{Finnish}} \\ \cmidrule(r){2-4} \cmidrule(r){5-7} \cmidrule(r){8-10} \cmidrule(r){11-13} \cmidrule(r){14-16} \cmidrule(r){17-19} \textbf{Class} & \textbf{Pr} & \textbf{Rec} & \textbf{F1} & \textbf{Pr} & \textbf{Rec} & \textbf{F1} & \textbf{Pr} & \textbf{Rec} & \textbf{F1} & \textbf{Pr} & \textbf{Rec} & \textbf{F1} & \textbf{Pr} & \textbf{Rec} & \textbf{F1} & \textbf{Pr} & \textbf{Rec} & \textbf{F1} \\ 
\midrule 
\sc Supportive & 0.43 & 0.33 & 0.38 & 0.37 & 0.26 & 0.30 & 0.44 & 0.34 & \bf 0.39 & 0.37 & 0.31 & 0.34 & 0.36 & 0.19 & 0.24 & 0.40 & 0.34 & 0.37 \\ 
\sc Indicator & 0.22 & 0.77 & 0.35 & 0.22 & 0.75 & 0.34 & 0.23 & 0.76 & \bf 0.36 & 0.20 & 0.64 & 0.31 & 0.22 & 0.79 & 0.34 & 0.23 & 0.74 & 0.35 \\ 
\sc Ideation & 0.43 & 0.19 & 0.27 & 0.42 & 0.20 & 0.27 & 0.40 & 0.22 & 0.28 & 0.41 & 0.22 & \bf 0.29 & 0.45 & 0.22 & \bf 0.29 & 0.41 & 0.21 & 0.28 \\ 
\sc Behavior & 1.00 & 0.01 & 0.03 & 0.67 & 0.03 & \bf 0.05 & 0.00 & 0.00 & 0.00 & 1.00 & 0.01 & 0.03 & 0.50 & 0.01 & 0.03 & 0.00 & 0.00 & 0.00 \\ 
\sc Attempt & 0.00 & 0.00 & 0.00 & 0.00 & 0.00 & 0.00 & 1.00 & 0.04 & \bf 0.09 & 0.00 & 0.00 & 0.00 & 0.50 & 0.02 & 0.04 & 1.00 & 0.04 & \bf 0.09 \\ 
\midrule \textbf{Macro avg} & 0.42 & 0.26 & 0.20 & 0.33 & 0.25 & 0.19 & 0.42 & 0.27 & \underline{\bf 0.22} & 0.40 & 0.24 & 0.19 & 0.40 & 0.24 & 0.19 & 0.41 & 0.27 & \underline{\bf 0.22} \\ 
\bottomrule \end{tabular} } \caption{\textbf{Back-translation and prediction with GPT-3.5-turbo with 0-shot learning on \textsc{C-ssrs}}, measuring Precision, Recall, and F1 per class per language. The last row shows the macro averages. The best F1 per class is shown in bold. The best average F1 across languages is underlined.} 
\label{results-suicide-gpt-3.5-back} 
\end{table*}

\paragraph{Further analysis with examples}

\begin{figure*}[t]
  \includegraphics[width=\textwidth]{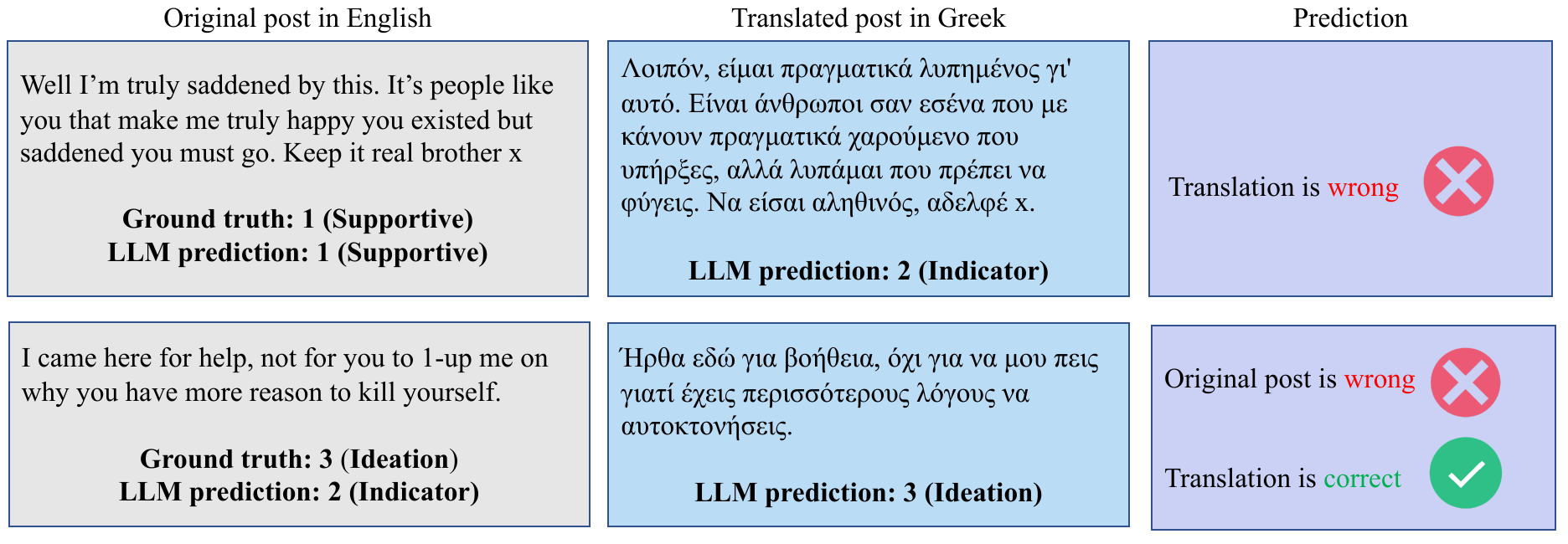}
  \caption{An example of translation (English to Greek) from the \textsc{C-ssrs} dataset with GPT-4o-mini.}
  \label{greek-suicide-example}
\end{figure*}

Lastly, Figure \ref{greek-suicide-example} presents two instances of the suicide dataset. 
Regarding the first example, while the translation from English to Greek appears to be correct in terms of conveying the same message, the English text is classified as "Supportive", while the Greek text is classified incorrectly as "Indicator".
In this case, GPT could interpret the context differently due to the nuances in the languages. 
In Greek, the model might be weighing other parts of the text more heavily, leading to a classification of "Indicator". 
This could be due to differences in the training data, the intricacies of the language model, or contextual emphasis.

For the second example, we observe that the translation is accurate and conveys the same message as the original English text. 
Nevertheless, the English text was incorrectly classified as level 2 (indicator).
However, the Greek translation was correctly classified as level 3 (ideation).
Here, we see that detecting different levels is inherently difficult.
The original English post could be mispredicted as it is quite small and generic.

\subsection{Resources and cost of experiments}
The total cost for running the GPT3.5 experiments via its API was less than \$30, showing the minimal resources required to conduct them, without the need for expensive GPU infrastructure or fine-tuning. 
For example, we experimented with MentaLLaMA-chat-7B \cite{yang2024mentallama}, which is a 27GB model. 
Its size makes it prohibitive when limited resources are available.
Eventually, we ended up using the quantized version, which did not output meaningful results. 
Our methodology for utilizing resources is potentially promising for extending the medical data sets (mostly in English) into other languages as well. 




\section{Discussion}
Despite the hype of using LLMs in mental healthcare for commercial applications, there is an increasing need for more research to understand the pitfalls and dangers of using LLMs across languages, diseases, and patient populations for mental health care. Considering the variation in the performance of the LLMs across languages, there is a need for utmost precautions not to rely on them solely in any healthcare setting. 
As stated by \newcite{stade2024large}, mental health diagnosis should never be left to automatic systems; human professionals are critical due to the possibility of errors and misdiagnoses, as illustrated above.

\noindent\textbf{LLMs exhibit considerable variability} in performance across languages, despite being evaluated on the same translated dataset. This inconsistency underscores the complexities inherent in multilingual mental health support, where language-specific nuances and mental health data coverage can effect the accuracy of the models. We observe performance gains in both tasks with stronger models and by translating to a specific target language (e.g., in German, for depression severity) before detection. On the other hand, translating to Turkish yields the worst performance in both tasks, which could be explained by the limited number of Turkish resources in health, especially in the domain of mental health \cite{ccoltekin2023resources}.

\section{Conclusion}
We presented a novel multilingual dataset designed to evaluate the ability of LLMs to predict the severity of mental health conditions in user-generated posts. Using English as the source language, we translated the posts to French, German, Turkish, Greek, Finnish, and Portuguese, using LLM-based machine translation. Then, we prompted the LLM to predict the severity of two mental health conditions across all languages. 
Our results indicate that the LLM performance varies depending on the language. 
Also, the best performance is not necessarily achieved in the source (English) language. 
Tested on low-resource languages (i.e., Greek and Turkish on medical domain), our method is useful for other languages due to its cost-effective nature.
Next steps comprise the application of our methodology to a broader range of languages (including a variety of low-resource languages) and adding datasets for more mental health tasks.



\section*{Limitations}

\paragraph{Translation} Creating a multilingual dataset requires significant resources.
In this work, we used a popular LLM like GPT3.5 to translate posts.
Translating using only an LLM and not having an expert or native-language human resources introduces a small loss of information that in some cases affects the final results.
Additionally, cultural differences are not ``transferred" correctly when using machine translation.

\paragraph{Evaluation} Automatically assessing the performance of LLMs is inherently challenging. To evaluate performance, we search for the assigned label within the LLM's output. If no label is detected, we assign the minimum label—class 0 for the depression dataset and class 2 for the suicide dataset (Indicator).

\paragraph{Potential risks} The quality of publicly available datasets is crucial, especially in sensitive areas such as mental health care. The datasets we utilized, as well as the multilingual data we created for this study, must be handled with great caution. These datasets should serve as tools to support healthcare professionals and their research, rather than being used to diagnose patients directly.

\bibliography{acl_latex}

@article{ccoltekin2023resources,
  title={Resources for Turkish natural language processing: A critical survey},
  author={{\c{C}}{\"o}ltekin, {\c{C}}a{\u{g}}r{\i} and Do{\u{g}}ru{\"o}z, A Seza and {\c{C}}etino{\u{g}}lu, {\"O}zlem},
  journal={Language Resources and Evaluation},
  volume={57},
  number={1},
  pages={449--488},
  year={2023},
  publisher={Springer}
}

@inproceedings{dogruoz-etal-2023-representativeness,
    title = "Representativeness as a Forgotten Lesson for Multilingual and Code-switched Data Collection and Preparation",
    author = {Do{\u{g}}ru{\"o}z, A. Seza  and
      Sitaram, Sunayana  and
      Yong, Zheng Xin},
    editor = "Bouamor, Houda  and
      Pino, Juan  and
      Bali, Kalika",
    booktitle = "Findings of the Association for Computational Linguistics: EMNLP 2023",
    month = dec,
    year = "2023",
    address = "Singapore",
    publisher = "Association for Computational Linguistics",
    url = "https://aclanthology.org/2023.findings-emnlp.382",
    doi = "10.18653/v1/2023.findings-emnlp.382",
    pages = "5751--5767",
}

@inproceedings{dogruoz-sitaram-2022-language,
    title = "Language Technologies for Low Resource Languages: Sociolinguistic and Multilingual Insights",
    author = {Do{\u{g}}ru{\"o}z, A. Seza  and
      Sitaram, Sunayana},
    editor = "Melero, Maite  and
      Sakti, Sakriani  and
      Soria, Claudia",
    booktitle = "Proceedings of the 1st Annual Meeting of the ELRA/ISCA Special Interest Group on Under-Resourced Languages",
    month = jun,
    year = "2022",
    address = "Marseille, France",
    publisher = "European Language Resources Association",
    url = "https://aclanthology.org/2022.sigul-1.12",
    pages = "92--97",
}

@article{stade2024large,
  title={Large language models could change the future of behavioral healthcare: a proposal for responsible development and evaluation},
  author={Stade, Elizabeth C and Stirman, Shannon Wiltsey and Ungar, Lyle H and Boland, Cody L and Schwartz, H Andrew and Yaden, David B and Sedoc, Jo{\~a}o and DeRubeis, Robert J and Willer, Robb and Eichstaedt, Johannes C},
  journal={NPJ Mental Health Research},
  volume={3},
  number={1},
  pages={12},
  year={2024},
  publisher={Nature Publishing Group UK London}
}

@article{brown2020,
  title={Language models are few-shot learners},
  author={Brown, Tom and Mann, Benjamin and Ryder, Nick and Subbiah, Melanie and Kaplan, Jared D and Dhariwal, Prafulla and Neelakantan, Arvind and Shyam, Pranav and Sastry, Girish and Askell, Amanda and others},
  journal={Advances in neural information processing systems},
  volume={33},
  pages={1877--1901},
  year={2020}
}

@inproceedings{devlin2019,
  title={BERT: Pre-training of deep bidirectional transformers for language understanding},
  author={Devlin, Jacob and Chang, Ming-Wei and Lee, Kenton and Toutanova, Kristina},
  booktitle={Proceedings of the 2019 Conference of the North American Chapter of the Association for Computational Linguistics: Human Language Technologies},
  year={2019}
}

@misc{who2021,
  title={Mental health},
  author={{World Health Organization}},
  year={2021},
  note={Retrieved from \url{https://www.who.int/health-topics/mental-health}}
}

@article{kessler2004,
  title={The World Mental Health (WMH) Survey Initiative Version of the World Health Organization (WHO) Composite International Diagnostic Interview (CIDI)},
  author={Kessler, Ronald C and Üstün, T Bedirhan},
  journal={International Journal of Methods in Psychiatric Research},
  volume={13},
  number={2},
  pages={93--121},
  year={2004},
  publisher={Wiley Online Library}
}

@article{guntuku2019,
  title={Studying expressions of loneliness in individuals using twitter: an observational study},
  author={Guntuku, Sharath Chandra and Schneider, Rachelle and Pelullo, Arthur and Young, Jami and Wong, Vivien and Ungar, Lyle and Polsky, Daniel and Volpp, Kevin G and Merchant, Raina},
  journal={BMJ open},
  volume={9},
  number={11},
  pages={e030355},
  year={2019},
  publisher={British Medical Journal Publishing Group}
}

@inproceedings{chancellor2019,
  title={A taxonomy of ethical tensions in inferring mental health states from social media},
  author={Chancellor, Stevie and Birnbaum, Michael L and Caine, Eric D and Silenzio, Vincent MB and De Choudhury, Munmun},
  booktitle={Proceedings of the Conference on Fairness, Accountability, and Transparency},
  pages={79--88},
  year={2019}
}

@article{lee2020biobert,
  title={BioBERT: a pre-trained biomedical language representation model for biomedical text mining},
  author={Lee, Jinhyuk and Yoon, Wonjin and Kim, Sungdong and Kim, Donghyeon and Kim, Sunkyu and So, Chan Ho and Kang, Jaewoo},
  journal={Bioinformatics},
  volume={36},
  number={4},
  pages={1234--1240},
  year={2020},
  publisher={Oxford University Press}
}

@article{clinicalbert,
author = {Kexin Huang and Jaan Altosaar and Rajesh Ranganath},
title = {ClinicalBERT: Modeling Clinical Notes and Predicting Hospital Readmission},
year = {2019},
journal = {arXiv:1904.05342},
}

@inproceedings{ji2022mentalbert,
  title={Mentalbert: Publicly available pretrained language models for mental healthcare},
  author={Ji, Shaoxiong and Zhang, Tianlin and Ansari, Luna and Fu, Jie and Tiwari, Prayag and Cambria, Erik},
  booktitle={proceedings of the thirteenth language resources and evaluation conference},
  pages={7184--7190},
  year={2022}
}

@inproceedings{aragon2023disorbert,
  title={DisorBERT: A double domain adaptation model for detecting signs of mental disorders in social media},
  author={Arag{\'o}n, Mario Ezra and L{\'o}pez-Monroy, A Pastor and Gonz{\'a}lez, Luis C and Losada, David E and Montes, Manuel},
  booktitle={Proceedings of the 61st Annual Meeting of the Association for Computational Linguistics (Volume 1: Long Papers)},
  pages={15305--15318},
  year={2023}
}

@inproceedings{raihan2024mentalhelp,
  title={MentalHelp: A Multi-Task Dataset for Mental Health in Social Media},
  author={Raihan, Nishat and Puspo, Sadiya Sayara Chowdhury and Farabi, Shafkat and Bucur, Ana-Maria and Ranasinghe, Tharindu and Zampieri, Marcos},
  booktitle={Proceedings of the 2024 Joint International Conference on Computational Linguistics, Language Resources and Evaluation (LREC-COLING 2024)},
  pages={11196--11203},
  year={2024}
}

@inproceedings{naseem2022early,
  title={Early identification of depression severity levels on reddit using ordinal classification},
  author={Naseem, Usman and Dunn, Adam G and Kim, Jinman and Khushi, Matloob},
  booktitle={Proceedings of the ACM Web Conference 2022},
  pages={2563--2572},
  year={2022}
}

@inproceedings{conneau-etal-2020-unsupervised,
    title = "Unsupervised Cross-lingual Representation Learning at Scale",
    author = "Conneau, Alexis  and
      Khandelwal, Kartikay  and
      Goyal, Naman  and
      Chaudhary, Vishrav  and
      Wenzek, Guillaume  and
      Guzm{\'a}n, Francisco  and
      Grave, Edouard  and
      Ott, Myle  and
      Zettlemoyer, Luke  and
      Stoyanov, Veselin",
    editor = "Jurafsky, Dan  and
      Chai, Joyce  and
      Schluter, Natalie  and
      Tetreault, Joel",
    booktitle = "Proceedings of the 58th Annual Meeting of the Association for Computational Linguistics",
    month = jul,
    year = "2020",
    address = "Online",
    publisher = "Association for Computational Linguistics",
    url = "https://aclanthology.org/2020.acl-main.747",
    doi = "10.18653/v1/2020.acl-main.747"
}

@inproceedings{xue-etal-2021-mt5,
    title = "m{T}5: A Massively Multilingual Pre-trained Text-to-Text Transformer",
    author = "Xue, Linting  and
      Constant, Noah  and
      Roberts, Adam  and
      Kale, Mihir  and
      Al-Rfou, Rami  and
      Siddhant, Aditya  and
      Barua, Aditya  and
      Raffel, Colin",
    editor = "Toutanova, Kristina  and
      Rumshisky, Anna  and
      Zettlemoyer, Luke  and
      Hakkani-Tur, Dilek  and
      Beltagy, Iz  and
      Bethard, Steven  and
      Cotterell, Ryan  and
      Chakraborty, Tanmoy  and
      Zhou, Yichao",
    booktitle = "Proceedings of the 2021 Conference of the North American Chapter of the Association for Computational Linguistics: Human Language Technologies",
    month = jun,
    year = "2021",
    address = "Online",
    publisher = "Association for Computational Linguistics",
    url = "https://aclanthology.org/2021.naacl-main.41",
    doi = "10.18653/v1/2021.naacl-main.41"
}

@inproceedings{zhang-etal-2020-improving,
    title = "Improving Massively Multilingual Neural Machine Translation and Zero-Shot Translation",
    author = "Zhang, Biao  and
      Williams, Philip  and
      Titov, Ivan  and
      Sennrich, Rico",
    editor = "Jurafsky, Dan  and
      Chai, Joyce  and
      Schluter, Natalie  and
      Tetreault, Joel",
    booktitle = "Proceedings of the 58th Annual Meeting of the Association for Computational Linguistics",
    month = jul,
    year = "2020",
    address = "Online",
    publisher = "Association for Computational Linguistics",
    url = "https://aclanthology.org/2020.acl-main.148",
    doi = "10.18653/v1/2020.acl-main.148",
}

@inproceedings{shing2018expert,
  title={Expert, crowdsourced, and machine assessment of suicide risk via online postings},
  author={Shing, Han-Chin and Nair, Suraj and Zirikly, Ayah and Friedenberg, Meir and Daum{\'e} III, Hal and Resnik, Philip},
  booktitle={Proceedings of the fifth workshop on computational linguistics and clinical psychology: from keyboard to clinic},
  pages={25--36},
  year={2018}
}

@inproceedings{coppersmith2014quantifying,
  title={Quantifying mental health signals in Twitter},
  author={Coppersmith, Glen and Dredze, Mark and Harman, Craig},
  booktitle={Proceedings of the workshop on computational linguistics and clinical psychology: From linguistic signal to clinical reality},
  pages={51--60},
  year={2014}
}

@inproceedings{coppersmith2015clpsych,
  title={CLPsych 2015 shared task: Depression and PTSD on Twitter},
  author={Coppersmith, Glen and Dredze, Mark and Harman, Craig and Hollingshead, Kristy and Mitchell, Margaret},
  booktitle={Proceedings of the 2nd workshop on computational linguistics and clinical psychology: from linguistic signal to clinical reality},
  pages={31--39},
  year={2015}
}

@inproceedings{de2014mental,
  title={Mental health discourse on reddit: Self-disclosure, social support, and anonymity},
  author={De Choudhury, Munmun and De, Sushovan},
  booktitle={Proceedings of the international AAAI conference on web and social media},
  volume={8},
  number={1},
  pages={71--80},
  year={2014}
}

@article{de2023benefits,
  title={Benefits and harms of large language models in digital mental health},
  author={De Choudhury, Munmun and Pendse, Sachin R and Kumar, Neha},
  journal={arXiv preprint arXiv:2311.14693},
  year={2023}
}

@article{ji2023domain,
  title={Domain-specific continued pretraining of language models for capturing long context in mental health},
  author={Ji, Shaoxiong and Zhang, Tianlin and Yang, Kailai and Ananiadou, Sophia and Cambria, Erik and Tiedemann, J{\"o}rg},
  journal={arXiv preprint arXiv:2304.10447},
  year={2023}
}

@article{harrigian2020state,
  title={On the state of social media data for mental health research},
  author={Harrigian, Keith and Aguirre, Carlos and Dredze, Mark},
  journal={arXiv preprint arXiv:2011.05233},
  year={2020}
}

@article{xu2024mental,
  title={Mental-llm: Leveraging large language models for mental health prediction via online text data},
  author={Xu, Xuhai and Yao, Bingsheng and Dong, Yuanzhe and Gabriel, Saadia and Yu, Hong and Hendler, James and Ghassemi, Marzyeh and Dey, Anind K and Wang, Dakuo},
  journal={Proceedings of the ACM on Interactive, Mobile, Wearable and Ubiquitous Technologies},
  volume={8},
  number={1},
  pages={1--32},
  year={2024},
  publisher={ACM New York, NY, USA}
}

@article{yang2023mentalllama,
  title={MentalLLaMA: Interpretable Mental Health Analysis on Social Media with Large Language Models},
  author={Yang, Kailai and Zhang, Tianlin and Kuang, Ziyan and Xie, Qianqian and Ananiadou, Sophia},
  journal={arXiv preprint arXiv:2309.13567},
  year={2023}
}

@inproceedings{yang2024mentallama,
  title={MentaLLaMA: interpretable mental health analysis on social media with large language models},
  author={Yang, Kailai and Zhang, Tianlin and Kuang, Ziyan and Xie, Qianqian and Huang, Jimin and Ananiadou, Sophia},
  booktitle={Proceedings of the ACM on Web Conference 2024},
  pages={4489--4500},
  year={2024}
}

@inproceedings{gaur2019knowledge,
  title={Knowledge-aware assessment of severity of suicide risk for early intervention},
  author={Gaur, Manas and Alambo, Amanuel and Sain, Joy Prakash and Kursuncu, Ugur and Thirunarayan, Krishnaprasad and Kavuluru, Ramakanth and Sheth, Amit and Welton, Randy and Pathak, Jyotishman},
  booktitle={The world wide web conference},
  pages={514--525},
  year={2019}
}

\clearpage

\appendix

\section{Appendix}
\label{sec:appendix}

Next, we present some examples of misprediction when using the original post, translated and back-translated text. All examples come from the \textsc{C-ssrs} dataset.

\subsection{Translation mispredicted, back-translation ok}

{\bf Original post:} ``Im sure that no ones lying to you. A man I know just succeeded in committing Suicide and I can assure you that he was a ghost of himself, consumed by major Mental Depression for months. Everyone cares, and everyone wishes he hadnt done it. I hope you can find a way to carry on, and I hope that if you give it some time you can find healing and happiness in your life." \\

\noindent {\bf Ground-truth:} 2. Indicator \\

\noindent {\bf Prediction:} 2. Indicator \\

\noindent Here the original post was correctly classified as ``2. Indicator". \\

\noindent {\bf Finnish translation:} ``Olen varma, ettei kukaan valehtele sinulle. Mies, jonka tunnen, onnistui juuri tekemään itsemurhan, ja voin vakuuttaa, että hän oli varjo entisestään, syötyään pahoista mielenterveysongelmista kuukausia. Kaikki välittävät, ja kaikki toivovat, ettei hän olisi tehnyt sitä. Toivon, että löydät tavan jatkaa eteenpäin, ja toivon, että jos annat sille aikaa, voit löytää parantumista ja onnea elämääsi." \\

\noindent {\bf Prediction}: 3. Ideation \\

\noindent {\bf Back-translation to English:} ``I am sure that no one is lying to you. The man I know just managed to commit suicide, and I can assure you that he was a shadow of his former self, having suffered from severe mental health issues for months. Everyone cares, and everyone wishes he hadn't done it. I hope you find a way to move forward, and I hope that if you give it time, you can find healing and happiness in your life." \\

\noindent {\bf Prediction with back-translation:} 2. Indicator \\

\noindent In this example we observe that the back-translation helped to correct the prediction.

\subsection{Translation ok, back-translation misprediction}

{\bf Original post:} ``Dont feel guilty. Better than doing nothing, get someone(his family or a friend) to show him that he is cared for. If you get a chance you could go personally and help him. For now keep him encouraged. Try to cheer him up and distract him from these thoughts. I understand how you feel. I also get such thoughts. If you need a friend, Ill be there for you(you can PM me). Dont waste yourself. You are a blessing to this world.Your family will feel Tired sad. BTW, its better you try taking that anti-depressing medication. Please throw away the gun. True. Everyone has a guardian angel."

\noindent {\bf Ground-truth:} 1. Supportive \\ \\
Prediction: 2. Indicator \\

\noindent Here the original post was misclassified as ``2. Indicator" while it belongs to ``1. Supportive". \\

\noindent {\bf Greek translation:} ``Μην νιώθεις ένοχος. Καλύτερα από το να μην κάνεις τίποτα, βρες κάποιον (την οικογένειά του ή ένα φίλο) να του δείξει ότι τον νοιάζεσαι. Αν έχεις την ευκαιρία, μπορείς να πας προσωπικά και να τον βοηθήσεις. Για τώρα, κράτα τον ενθαρρυμένο. Προσπάθησε να τον ανεβάσεις τα ηθικά και να τον αποσπάσεις από αυτές τις σκέψεις. Καταλαβαίνω πώς νιώθεις. Κι εγώ έχω τέτοιες σκέψεις. Αν χρειάζεσαι ένα φίλο, θα είμαι εδώ για σένα (μπορείς να μου στείλεις μήνυμα). Μην σπαταλάς τον εαυτό σου. Είσαι μια ευλογία για αυτόν τον κόσμο. Η οικογένειά σου θα νιώσει κουρασμένη και λυπημένη. Επίσης, είναι καλύτερα να δοκιμάσεις να πάρεις αυτό το αντικαταθλιπτικό φάρμακο. Παρακαλώ πέταξε μακριά το όπλο. Αληθινό. Κάθε άνθρωπος έχει έναν φύλακα άγγελο." \\

\noindent {\bf Prediction with Greek translation:} 1. Supportive \\

\noindent{\bf Back-translation to English}: ``Do not feel guilty. Instead of doing nothing, find someone (their family or a friend) to show them that you care. If you have the opportunity, you can go in person and help them. For now, keep them encouraged. Try to lift their spirits and distract them from those thoughts. I understand how you feel. I also have such thoughts. If you need a friend, I will be here for you (you can message me). Do not waste yourself. You are a blessing to this world. Your family will feel tired and sad. Also, it is better to try taking this antidepressant medication. Please throw the gun away. True. Every person has a guardian angel." \\

\noindent {\bf Prediction with back-translated:} 2. Indicator. \\

\noindent We observe that the text was incorrectly predicted, while the Greek translation was predicted correctly.

\end{document}